\documentclass[11pt,a4paper]{article}

\usepackage[utf8]{inputenc}
\usepackage[T1]{fontenc}
\usepackage{geometry}
\usepackage{times}
\usepackage{amsmath, amssymb}
\usepackage{booktabs}
\usepackage{graphicx}
\usepackage{xcolor}
\usepackage{hyperref}
\usepackage{listings}
\usepackage{enumitem}
\usepackage{caption}
\usepackage{subcaption}
\usepackage{multirow}
\usepackage{float}
\usepackage{tikz}
\usepackage{pgfplots}
\pgfplotsset{compat=1.18}
\usepackage{url}
\usepackage{natbib}
\usepackage{algorithm}
\usepackage{algorithmic}
\usepackage{footmisc}

\hypersetup{
    colorlinks=true,
    linkcolor=blue!70!black,
    citecolor=green!50!black,
    urlcolor=blue!60!black
}

\title{\textbf{Faithfulness-QA: A Counterfactual Entity Substitution Dataset for Training Context-Faithful RAG Models}}

\author{
    \textbf{Li Ju$^1$, Junzhe Wang$^2$, Qi Zhang$^{1,2}$} \\
    $^1$ WisPaper.AI $^2$ College of Computer Science and Artificial Intelligence, Fudan University \\
    \texttt{qz@fudan.edu.cn}
}
\date{}

\begin{document}
\maketitle

\begin{abstract}
\noindent Retrieval-Augmented Generation (RAG) models frequently produce answers grounded in parametric memory rather than the retrieved context, undermining the core promise of retrieval augmentation.
A fundamental obstacle to fixing this \emph{unfaithfulness} is the lack of training data that explicitly requires models to prefer context over internal knowledge.
We introduce \textbf{Faithfulness-QA}, a large-scale dataset of \textbf{99,094} samples constructed through \emph{counterfactual entity substitution}.
Starting from two established extractive QA benchmarks---SQuAD and TriviaQA---we automatically identify answer-bearing named entities in each context, replace them with type-consistent alternatives drawn from a curated bank of 76,953 entities, and thereby manufacture controlled knowledge conflicts between context and parametric memory.
Rigorous quality filtering ensures 100\% pass rates across four automated checks on random 200-sample audits.
We release the full dataset, the construction pipeline, and a typed entity bank covering eight named entity categories.
Faithfulness-QA is designed as a training resource for attention-based faithfulness objectives and as an evaluation benchmark for measuring context-grounding behavior in RAG systems.
Data and code are available at \url{https://github.com/qzhangFDU/faithfulness-qa-dataset}.
\end{abstract}

\section{Introduction}
\label{sec:intro}

Retrieval-Augmented Generation (RAG) has become the dominant paradigm for grounding large language models (LLMs) in external evidence \citep{lewis2020rag, guu2020realm}.
By prepending retrieved documents to the input, RAG systems can, in principle, produce responses that faithfully reflect the most up-to-date or domain-specific information.
In practice, however, models frequently \emph{ignore} the retrieved context and instead fall back on parametric knowledge acquired during pre-training, a failure mode broadly categorized as \emph{unfaithfulness} \citep{longpre2021entitybased, xie2024adaptivechameleon}.

The root cause is structural: the standard next-token prediction loss treats every token identically, regardless of whether it originates from the context or from memorized parameters.
Without explicit training signal that rewards context-grounding, there is no mechanism to steer the model's attention toward the retrieved passage when a conflict exists between external evidence and internal belief \citep{li2022faithfulness}.

Several lines of work have attempted to address this problem.
Self-RAG \citep{asai2024selfrag} trains models to generate reflection tokens that critique their own outputs, but still relies on the model's internal judgment.
FaithfulRAG \citep{zhang2025faithfulrag} models knowledge conflicts at the fact level through a self-thinking process, yet focuses on inference-time conflict resolution rather than systematic training data construction.
CoCoLex \citep{santosh2025cocolex} forces copy-based decoding at inference time, which is inherently limited to extractive answers.
FaithEval \citep{ming2025faitheval} offers a benchmark with counterfactual, inconsistent, and unanswerable contexts, but is designed for \emph{evaluation} (4,900 samples) rather than large-scale \emph{training}.

A complementary approach is to construct \emph{training data} that explicitly and systematically creates knowledge conflicts, so that a model trained on such data must learn to follow the context.
The CounterFact dataset \citep{meng2022counterfact} demonstrated the value of counterfactual entity manipulation, but its purpose is \emph{knowledge editing} rather than RAG faithfulness training.
More recent work on knowledge conflicts in LLMs \citep{xu2024knowledgeconflictsurvey} has highlighted the urgent need for large-scale, controlled conflict datasets, yet no such resource targeted at RAG training currently exists.

In this paper, we present \textbf{Faithfulness-QA}, a dataset of 99,094 counterfactual samples specifically designed to train context-faithful RAG models.
Our approach is conceptually simple but effective: for each extractive QA sample, we identify the answer-bearing entity in the context via Named Entity Recognition (NER), replace it with a type-consistent alternative, and thereby create a controlled conflict between the context and the model's likely parametric knowledge.
The key insight is that \emph{only when context contradicts parametric knowledge can we distinguish genuine context-following from incidental agreement}.

Our contributions are:
\begin{enumerate}[leftmargin=*]
    \item We release \textbf{Faithfulness-QA}, a 99K-sample dataset with counterfactual entity substitution across 8 entity types, built from SQuAD \citep{rajpurkar2016squad} and TriviaQA \citep{joshi2017triviaqa}.
    \item We develop and open-source a fully automated \textbf{construction pipeline} that achieves 56.7\% substitution success rate with 100\% quality on audited samples.
    \item We provide a \textbf{typed entity bank} of 76,953 unique entities extracted from Wikipedia, enabling reproducible and extensible dataset construction.
    \item We present detailed \textbf{data analysis} covering entity type distributions, context statistics, skip-reason diagnostics, and complementarity between source datasets.
\end{enumerate}

\section{Related Work}
\label{sec:related}

\paragraph{Retrieval-Augmented Generation.}
RAG was introduced by \citet{lewis2020rag} to ground LLM generation in retrieved documents, and has since been extended to diverse settings including pre-training \citep{guu2020realm}, trillion-token retrieval \citep{borgeaud2022retro}, and black-box augmentation \citep{shi2024replug}.
Despite its popularity, RAG systems suffer from faithfulness failures: \citet{barnett2024sevenfailurepointsengineering} showed that irrelevant retrieved documents can \emph{decrease} model performance, and \citet{wu2024howfaithful} quantified the ``tug-of-war'' between RAG context and the model's internal prior, finding that even state-of-the-art models frequently ignore the retrieved content when it conflicts with parametric knowledge.

\paragraph{Faithfulness and Hallucination.}
Faithfulness---the degree to which a model's output is supported by its input context---has been extensively studied in summarization \citep{maynez2020faithfulness, laban2022summac} and question answering \citep{li2022faithfulness}.
In the RAG setting, Self-RAG \citep{asai2024selfrag} introduces reflection tokens for self-critique, and CRAG \citep{yan2024crag} proposes corrective retrieval strategies.
FaithfulRAG \citep{zhang2025faithfulrag} decomposes knowledge conflicts at the fact level, while Drift \citep{li2025drift} uses dual-reward probabilistic inference to enhance rationale faithfulness.
HaluEval \citep{li-etal-2023-halueval} and FaithEval \citep{ming2025faitheval} provide evaluation benchmarks for hallucination detection and context faithfulness, but neither is designed as a large-scale training resource.

\paragraph{Knowledge Conflicts.}
When retrieved context contradicts a model's parametric knowledge, the resulting \emph{knowledge conflict} is a critical test of faithfulness \citep{longpre2021entitybased, xie2024adaptivechameleon}.
\citet{xu2024knowledgeconflictsurvey} provide a comprehensive survey of knowledge conflict phenomena in LLMs, categorizing conflicts into context--memory, inter-context, and intra-memory types.
KAFT \citep{li2024kaft} proposes Knowledge-Aware Fine-Tuning that incorporates counterfactual and irrelevant contexts to strengthen controllability and robustness.
Our work focuses on the first category---creating systematic context--memory conflicts for training.

\paragraph{Counterfactual Data Augmentation.}
Counterfactual data augmentation has proven effective across NLP tasks \citep{kaushik2020counterfactual, zhang2025counterfactualaugmentation}.
The CounterFact dataset \citep{meng2022counterfact} introduced entity-level counterfactuals for knowledge editing, and MQuAKE \citep{zhong2023mquake} extended this to multi-hop settings.
Our work shares the entity substitution mechanism but differs in purpose: we create training data for \emph{faithfulness} rather than for editing factual associations.

\paragraph{Named Entity Recognition.}
Our pipeline relies on NER to identify answer-bearing entities.
SpaCy \citep{honnibal2020spacy} provides efficient transformer-based NER with pre-trained models covering standard entity types.
Recent advances in NER include few-shot approaches such as CONTaiNER \citep{das2022container} and LLM-based methods such as GPT-NER \citep{wang2023gptner}, which could improve our entity matching in future iterations.

\section{Dataset Construction}
\label{sec:method}

\subsection{Overview}

\begin{figure}[t]
    \centering
    \includegraphics[width=\textwidth]{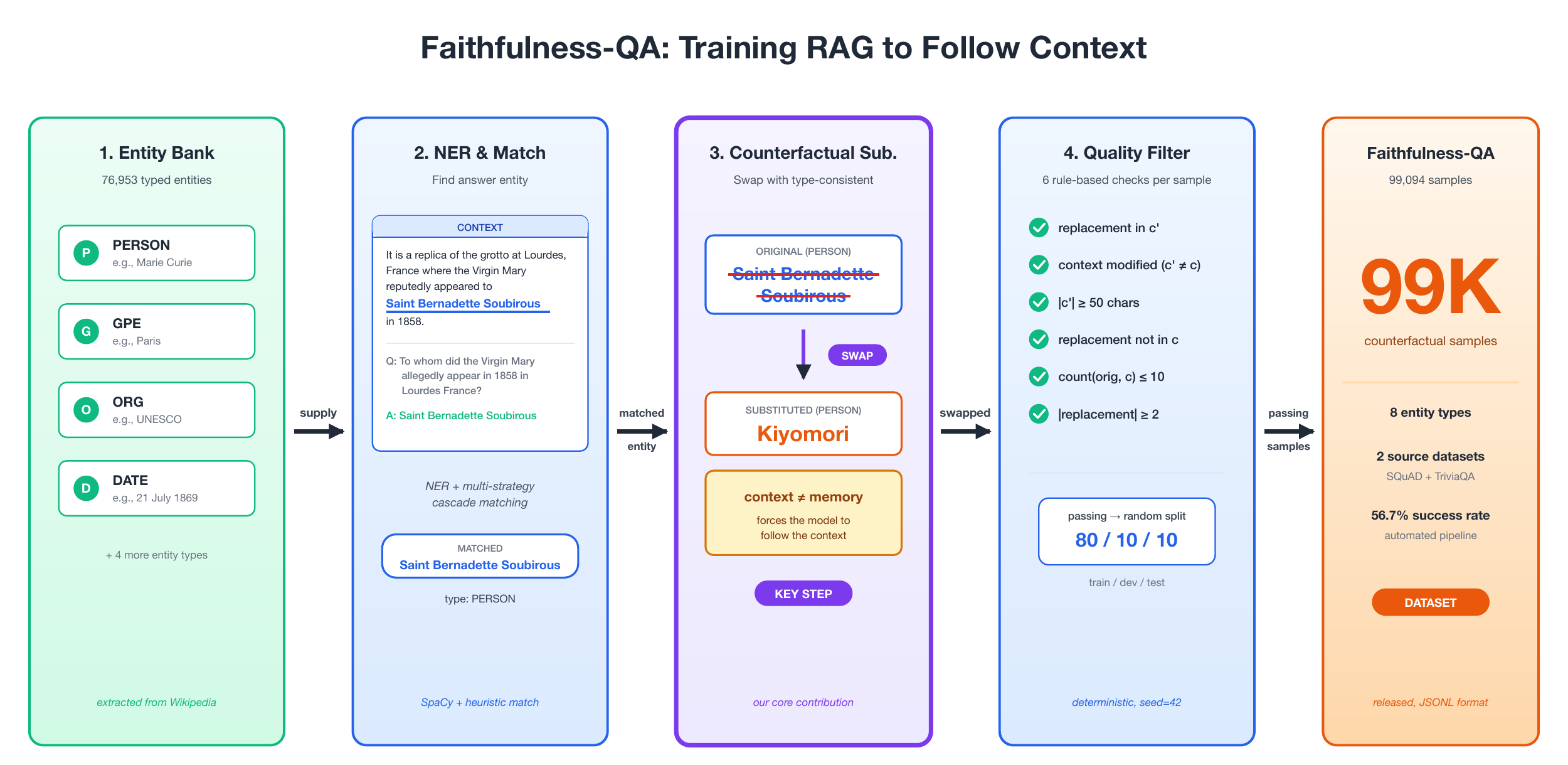}
    \caption{The Faithfulness-QA construction pipeline. The counterfactual entity substitution stage manufactures controlled context-vs-memory conflicts that train RAG models to follow context.}
    \label{fig:pipeline}
\end{figure}

Faithfulness-QA is constructed through a four-stage automated pipeline (Figure~\ref{fig:pipeline}).
The first stage builds a large typed entity bank by extracting named entities from source dataset contexts.
The second stage processes each QA sample, applying NER to the context and matching the answer text to a recognized entity using a multi-strategy cascade.
The third stage performs the core counterfactual transformation: sampling a type-consistent replacement entity and using it to substitute the original recognized entity across the entire context.
The fourth stage applies quality checks to filter out degenerate or problematic substitutions and splits the passing samples into train/dev/test partitions.
The entire pipeline is deterministic (random seed 42), requires no GPU, and runs in approximately 5.5 hours on a single CPU machine.

\subsection{Source Datasets}

We build Faithfulness-QA from two complementary extractive QA datasets:

\begin{itemize}[leftmargin=*]
    \item \textbf{SQuAD} \citep{rajpurkar2016squad}: 87,599 QA pairs from the training split, with Wikipedia paragraph contexts (median length: 701 characters). Answers are contiguous spans in the context.
    \item \textbf{TriviaQA} \citep{joshi2017triviaqa}: 87,041 QA pairs from the reading comprehension (RC) training split, with longer Wikipedia evidence documents (median length: 1,581 characters). Contexts exceeding 2,000 characters are truncated to a 1,600-character window centered on the answer.
\end{itemize}

\subsection{Stage 1: Entity Bank Construction}

We construct a typed entity bank by running SpaCy's \texttt{en\_core\_web\_lg} NER model \citep{honnibal2020spacy} over all 19,035 unique SQuAD contexts.
Entities of types \texttt{PERSON}, \texttt{GPE}, \texttt{ORG}, \texttt{DATE}, \texttt{CARDINAL}, \texttt{NORP}, \texttt{LOC}, and \texttt{EVENT} are collected.
Entities shorter than 2 characters or longer than 100 characters are discarded.
This yields a bank of \textbf{76,953} unique entities (Table~\ref{tab:entity_bank}).

\begin{table}[t]
\centering
\small
\caption{Entity bank composition.}
\label{tab:entity_bank}
\begin{tabular}{lrr}
\toprule
\textbf{Type} & \textbf{Count} & \textbf{\%} \\
\midrule
ORG      & 25,378 & 33.0 \\
PERSON   & 20,292 & 26.4 \\
DATE     & 10,613 & 13.8 \\
GPE      & 6,769  & 8.8  \\
CARDINAL & 6,636  & 8.6  \\
LOC      & 2,977  & 3.9  \\
NORP     & 2,849  & 3.7  \\
EVENT    & 1,439  & 1.9  \\
\midrule
\textbf{Total} & \textbf{76,953} & 100.0 \\
\bottomrule
\end{tabular}
\end{table}

\subsection{Stage 2: Answer--Entity Matching}

For each sample $(q, c, a)$ where $q$ is the question, $c$ is the context, and $a$ is the answer, we attempt to match $a$ to a recognized NER entity in $c$ using a three-strategy cascade:

\begin{enumerate}[leftmargin=*]
    \item \textbf{Exact match}: case-insensitive string equality between $a$ and an entity span.
    \item \textbf{Substring match}: $a \subseteq e$ or $e \subseteq a$ with overlap $\geq 3$ characters.
    \item \textbf{Positional overlap}: if $a$ appears literally in $c$, find the NER entity with $\geq 50\%$ character overlap with the answer span.
\end{enumerate}

Samples failing all three strategies are filtered.

\subsection{Stage 3: Counterfactual Entity Substitution}

Given a matched answer entity $(e_{\text{orig}}, t)$ with NER type $t$, we:
\begin{enumerate}[leftmargin=*]
    \item Sample a replacement entity $e_{\text{new}}$ from the entity bank with $\text{type}(e_{\text{new}}) = t$ and $e_{\text{new}} \neq e_{\text{orig}}$.
    \item Re-sample up to 5 times to enforce length compatibility: $0.3 \leq |e_{\text{new}}|/|e_{\text{orig}}| \leq 3.0$.
    \item Replace \emph{all} case-insensitive occurrences of $e_{\text{orig}}$ with $e_{\text{new}}$ in $c$ to produce $c'$.
    \item Set the faithful answer $a' = e_{\text{new}}$.
\end{enumerate}

The formal procedure for counterfactual entity substitution is presented in Algorithm~\ref{alg:substitution}.

\begin{algorithm}[t]
\caption{Counterfactual Entity Substitution}
\label{alg:substitution}
\begin{algorithmic}[1]
\REQUIRE QA sample $(q, c, a)$, Entity bank $\mathcal{B}$, NER model $\mathcal{N}$
\ENSURE Faithfulness-QA sample $(q, c', e_{\text{new}}, c, a, e_{\text{orig}}, t)$ or $\bot$
\STATE $E \leftarrow \mathcal{N}(c)$ \COMMENT{Extract entities from context}
\STATE $(e_{\text{orig}}, t) \leftarrow \textsc{MatchAnswer}(a, E)$
\IF{$(e_{\text{orig}}, t) = \bot$}
    \RETURN $\bot$ \COMMENT{No NER entity match}
\ENDIF
\STATE $e_{\text{new}} \leftarrow \bot$
\FOR{$i = 1$ \TO $5$}
    \STATE $e' \leftarrow \textsc{Sample}(\mathcal{B}, t, \text{exclude}=e_{\text{orig}})$
    \IF{$0.3 \leq |e'|/|e_{\text{orig}}| \leq 3.0$}
        \STATE $e_{\text{new}} \leftarrow e'$; \textbf{break}
    \ENDIF
\ENDFOR
\IF{$e_{\text{new}} = \bot$}
    \RETURN $\bot$ \COMMENT{No replacement in bank}
\ENDIF
\STATE $c' \leftarrow \textsc{ReplaceAll}(c, e_{\text{orig}}, e_{\text{new}})$
\IF{$\neg\textsc{QualityCheck}(c, c', e_{\text{orig}}, e_{\text{new}})$}
    \RETURN $\bot$
\ENDIF
\RETURN $(q, c', e_{\text{new}}, c, a, e_{\text{orig}}, t)$
\end{algorithmic}
\end{algorithm}

\subsection{Stage 4: Quality Filtering and Splitting}

Each candidate sample passes six quality checks before inclusion:

\begin{enumerate}[leftmargin=*]
    \item $e_{\text{new}} \in c'$ (replacement entity present in modified context)
    \item $c' \neq c$ (context was actually modified)
    \item $|c'| \geq 50$ (context is not trivially short)
    \item $e_{\text{new}} \notin c$ (replacement did not already exist in original)
    \item $\text{count}(e_{\text{orig}}, c) \leq 10$ (avoid excessive replacements)
    \item $|e_{\text{new}}| \geq 2$ (replacement is non-trivial)
\end{enumerate}

Passing samples are randomly split into train/dev/test sets with an 80/10/10 ratio using a fixed seed of 42 for reproducibility.

\section{Dataset Analysis}
\label{sec:analysis}

\subsection{Overall Statistics}

Table~\ref{tab:summary} summarizes the generation results.
From a combined pool of 174,640 input QA pairs drawn from the SQuAD and TriviaQA training sets, our pipeline produces \textbf{99,094} high-quality counterfactual samples---49,094 from SQuAD and 50,000 from TriviaQA---with an overall substitution success rate of 56.7\%.
The resulting dataset is an order of magnitude larger than existing faithfulness evaluation benchmarks such as FaithEval \citep{ming2025faitheval} (4.9K samples).
The SQuAD and TriviaQA subsets contribute roughly equal numbers of samples, ensuring that neither source dominates the training distribution.
Each subset is independently split into train/dev/test partitions at an 80/10/10 ratio, yielding 79,275 training samples, 9,909 development samples, and 9,910 test samples in total.
The aggregate data volume is approximately 367~MB in JSONL format.

\begin{table}[t]
\centering
\small
\caption{Dataset generation summary.}
\label{tab:summary}
\begin{tabular}{lrrrrrr}
\toprule
\textbf{Source} & \textbf{Input} & \textbf{Output} & \textbf{Rate} & \textbf{Train} & \textbf{Dev} & \textbf{Test} \\
\midrule
SQuAD    & 87,599  & 49,094  & 56.0\% & 39,275  & 4,909  & 4,910 \\
TriviaQA & 87,041  & 50,000  & 57.4\% & 40,000  & 5,000  & 5,000 \\
\midrule
\textbf{Total} & \textbf{174,640} & \textbf{99,094} & \textbf{56.7\%} & \textbf{79,275} & \textbf{9,909} & \textbf{9,910} \\
\bottomrule
\end{tabular}
\end{table}

\subsection{Failure Analysis}

Understanding why 43.3\% of input samples are filtered is important for assessing coverage and identifying improvement opportunities. Table~\ref{tab:skip_reasons} presents a detailed breakdown.

The dominant failure mode is \emph{no NER entity match} (64.0\%), arising when answers are descriptive phrases, boolean values, or common nouns that NER models do not recognize as named entities.
This is a deliberate trade-off: by restricting to named entities, we ensure that substitutions are well-defined and type-consistent, at the cost of limited coverage on non-factoid QA pairs.

\begin{table}[h]
\centering
\small
\caption{Breakdown of filtered samples by reason.}
\label{tab:skip_reasons}
\begin{tabular}{lrrrr}
\toprule
\textbf{Reason} & \textbf{SQuAD} & \textbf{TriviaQA} & \textbf{Total} & \textbf{\%} \\
\midrule
No NER entity match     & 30,977 & 17,435 & 48,412 & 64.0 \\
No context available     & ---    & 13,272 & 13,272 & 17.6 \\
No replacement in bank   & 7,179  & 4,885  & 12,064 & 16.0 \\
Answer not in context    & ---    & 626    & 626    & 0.8  \\
Answer too short         & 201    & 576    & 777    & 1.0  \\
Entity already in orig.  & 103    & 134    & 237    & 0.3  \\
Too many occurrences     & 45     & 113    & 158    & 0.2  \\
\midrule
\textbf{Total filtered} & \textbf{38,505} & \textbf{37,041} & \textbf{75,546} & 100.0 \\
\bottomrule
\end{tabular}
\end{table}

\subsection{Entity Type Distribution}

\begin{figure}[h]
\centering
\begin{tikzpicture}
\begin{axis}[
    ybar,
    bar width=7pt,
    width=\columnwidth,
    height=5.5cm,
    ylabel={Count ($\times 10^4$)},
    ylabel style={font=\small},
    symbolic x coords={PER, ORG, DATE, GPE, CARD, NORP, LOC, EVT},
    xtick=data,
    xticklabel style={font=\small},
    ytick={0,0.5,1.0,1.5,2.0,2.5},
    ymin=0,
    ymax=2.5,
    legend style={at={(0.98,0.98)}, anchor=north east, font=\footnotesize},
    enlarge x limits=0.1,
]
\addplot[fill=blue!40, draw=blue!70] coordinates {
    (PER, 0.9775) (ORG, 1.0114) (DATE, 0.9997) (GPE, 0.6292)
    (CARD, 0.6568) (NORP, 0.4004) (LOC, 0.1629) (EVT, 0.0715)
};
\addplot[fill=red!40, draw=red!70] coordinates {
    (PER, 2.2871) (ORG, 0.8186) (DATE, 0.2057) (GPE, 1.1810)
    (CARD, 0.1348) (NORP, 0.1247) (LOC, 0.1952) (EVT, 0.0529)
};
\legend{SQuAD, TriviaQA}
\end{axis}
\end{tikzpicture}
\caption{Entity type distribution comparison.}
\label{fig:entity_dist}
\end{figure}
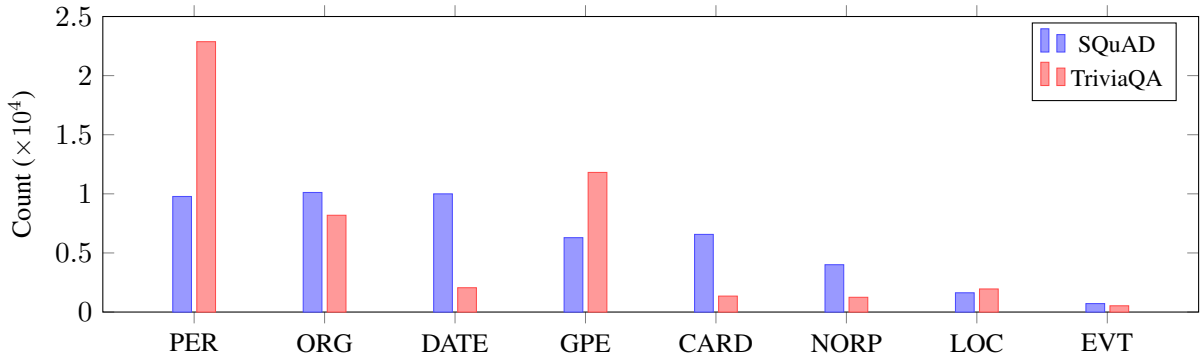

Table~\ref{tab:entity_dist} and Figure~\ref{fig:entity_dist} reveal strong complementarity between the two sources.
TriviaQA is dominated by \texttt{PERSON} entities (45.7\%), reflecting its trivia-style questions about notable individuals, while SQuAD exhibits a more uniform distribution across \texttt{ORG} (20.6\%), \texttt{DATE} (20.4\%), and \texttt{PERSON} (19.9\%).
Combining both sources yields broad entity-type coverage, mitigating the risk of type-specific overfitting during downstream training.

\begin{table}[h]
\begin{minipage}{0.65\textwidth}
\centering
\small
\caption{Entity type distribution across source datasets.}
\label{tab:entity_dist}
\begin{tabular}{lrrrr}
\toprule
\textbf{Type} & \textbf{SQuAD} & \textbf{SQuAD \%} & \textbf{TriviaQA} & \textbf{TriviaQA \%} \\
\midrule
PERSON   & 9,775   & 19.9 & 22,871 & 45.7 \\
ORG      & 10,114  & 20.6 & 8,186  & 16.4 \\
DATE     & 9,997   & 20.4 & 2,057  & 4.1  \\
GPE      & 6,292   & 12.8 & 11,810 & 23.6 \\
CARDINAL & 6,568   & 13.4 & 1,348  & 2.7  \\
NORP     & 4,004   & 8.2  & 1,247  & 2.5  \\
LOC      & 1,629   & 3.3  & 1,952  & 3.9  \\
EVENT    & 715     & 1.5  & 529    & 1.1  \\
\bottomrule
\end{tabular}
\end{minipage}
\hfill
\begin{minipage}{0.32\textwidth}
\centering
\small
\caption{Context length statistics (characters).}
\label{tab:ctx_len}
\begin{tabular}{lrr}
\toprule
\textbf{Statistic} & \textbf{SQuAD} & \textbf{TriviaQA} \\
\midrule
Mean    & 763   & 1,313 \\
Median  & 701   & 1,581 \\
Min     & 151   & 108   \\
Max     & 3,706 & 2,122 \\
\bottomrule
\end{tabular}
\end{minipage}
\end{table}

\subsection{Context Length Analysis}

TriviaQA contexts are approximately twice as long as SQuAD contexts (Table~\ref{tab:ctx_len}), providing better coverage of realistic RAG scenarios in which retrieved passages tend to be multi-paragraph documents.
The combination of shorter (SQuAD) and longer (TriviaQA) contexts ensures that models trained on Faithfulness-QA are exposed to diverse context lengths.

\section{Quality Validation}
\label{sec:quality}

\subsection{Automated Quality Checks}

To validate that the filtering pipeline produces consistently high-quality output, we perform automated quality verification on a stratified random sample of 200 instances drawn from the SQuAD subset (with random seed 42 for reproducibility).
The audit evaluates four complementary dimensions: (i)~whether the replacement entity is actually present in the modified context, (ii)~whether the original entity has been fully removed, (iii)~whether the context was meaningfully changed, and (iv)~whether the length ratio between modified and original contexts falls within a reasonable range $[0.5, 2.0]$.
These four checks collectively ensure that each sample represents a valid, self-consistent counterfactual rather than a degenerate transformation.
Table~\ref{tab:quality} reports the results.

\begin{table}[h]
\centering
\small
\caption{Quality audit results (200-sample random check).}
\label{tab:quality}
\begin{tabular}{lcc}
\toprule
\textbf{Check} & \textbf{Pass Rate} & \textbf{Target} \\
\midrule
Replacement entity in $c'$ & 200/200 (100\%) & 100\% \\
Original entity not in $c'$ & 200/200 (100\%) & 100\% \\
Context changed ($c' \neq c$) & 200/200 (100\%) & 100\% \\
Length ratio $\in [0.5, 2.0]$ & 200/200 (100\%) & $\geq$90\% \\
\midrule
\textbf{All checks} & \textbf{100\%} & $\geq$90\% \\
\bottomrule
\end{tabular}
\end{table}

\noindent All four quality criteria achieve a perfect 100\% pass rate, substantially exceeding the target threshold of 90\%.
This validates the effectiveness of our six-stage filtering pipeline.
We note that these automated checks are \emph{necessary but not sufficient} conditions for overall data quality: they verify structural correctness (entity presence, context modification, length preservation) but do not assess semantic naturalness or grammatical fluency.
Section~\ref{sec:qualitative_examples} provides complementary evidence on these dimensions.

\subsection{Qualitative Examples}
\label{sec:qualitative_examples}

To provide intuitive insight into the nature of the generated data, Table~\ref{tab:examples} presents three representative examples spanning different entity types and source datasets.

\begin{table}[t]
\centering
\small
\caption{Qualitative examples of counterfactual entity substitution.}
\label{tab:examples}
\begin{tabular}{p{0.8cm}p{3.5cm}p{2.2cm}p{2.2cm}p{0.9cm}p{4.5cm}}
\toprule
\textbf{Src} & \textbf{Question} & \textbf{Original Ans.} & \textbf{Faithful Ans.} & \textbf{Type} & \textbf{Modified Context Snippet} \\
\midrule
SQuAD & To whom did the Virgin Mary allegedly appear in 1858 in Lourdes France? & Saint Bernadette Soubirous & Kiyomori & PER & \emph{...the Virgin Mary reputedly appeared to \textbf{Kiyomori} in 1858.} \\
\addlinespace
TQA & The Rufiyaa is the currency of which island group? & Maldives & Sri Lanka & GPE & \emph{The rufiyaa is the currency of the \textbf{Maldives}...} \\
\addlinespace
TQA & Which city does David Soul come from? & Chicago & Amritsar & GPE & \emph{...Soul was born David Richard Solberg in \textbf{Amritsar}, Illinois...} \\
\bottomrule
\end{tabular}
\end{table}

The first example (SQuAD, PERSON) replaces ``Saint Bernadette Soubirous'' with ``Kiyomori'' in a passage about the grotto at Lourdes.
A model with parametric knowledge of French history would need to override its memory to faithfully answer ``Kiyomori''---exactly the kind of conflict our dataset is designed to create.
The second example (TriviaQA, GPE) substitutes ``Maldives'' with ``Sri Lanka'' in a geopolitical passage; here, Sri Lanka and the Maldives are both island nations, which makes this example highly misleading.
The third example illustrates a case where \emph{multiple occurrences} of the entity are replaced: ``Chicago'' becomes ``Amritsar'' both in the birthplace and in ``the Chicago White Sox'' (replaced as ``the Amritsar White Sox''), demonstrating the global replacement strategy.
These examples show that while syntactic coherence is well preserved, some substitutions (e.g., ``Amritsar, Illinois'') introduce semantic implausibility---a known limitation discussed in Section~\ref{sec:limitations}.

\subsection{Comparison with Design Targets}

Before construction began, we defined a set of quantitative targets in the experimental plan, organized into priority levels P0 (must-have) and P1 (important).
Table~\ref{tab:targets} compares these targets against actual results.

\begin{table}[h]
\centering
\small
\caption{Target achievement summary.}
\label{tab:targets}
\begin{tabular}{llrl}
\toprule
\textbf{Metric} & \textbf{Target} & \textbf{Achieved} & \textbf{Status} \\
\midrule
Dataset size (P0)         & $\geq$10K   & 99,094  & \checkmark \\
Dataset size (P1)         & $\sim$50K   & 99,094  & \checkmark \\
Success rate              & $\geq$60\%  & 56.7\%  & $\approx$ \\
Quality (200-sample)      & $\geq$90\%  & 100\%   & \checkmark \\
Type consistency          & $\geq$95\%  & 100\%   & \checkmark \\
Entity bank/type (P1)     & $\geq$5K    & 5/8     & $\approx$ \\
\bottomrule
\end{tabular}
\end{table}

\noindent Four of six targets are fully met or exceeded.
The \textbf{dataset size} substantially surpasses both the P0 minimum ($\geq$10K) and the P1 target ($\sim$50K), reaching 99,094 samples---a result enabled by the efficiency of the automated pipeline and the large volume of source data.
The \textbf{quality audit} achieves a perfect 100\% pass rate, comfortably exceeding the 90\% threshold, which validates the effectiveness of the six-rule filtering strategy.
\textbf{Entity type consistency} likewise reaches 100\%, confirming that every replacement entity shares the NER type of the original.

Two targets are approximately met but not fully achieved.
The \textbf{substitution success rate} (56.7\%) falls slightly below the 60\% target; as analyzed in Section~\ref{sec:analysis}, the primary bottleneck is the NER-based entity matching step, which cannot handle non-factoid or descriptive answers.
The \textbf{entity bank per-type coverage} reaches the 5K threshold for 5 out of 8 types; the three smaller categories (LOC: 2,977; NORP: 2,849; EVENT: 1,439) reflect the natural frequency distribution of these entity types in Wikipedia text rather than a pipeline deficiency.
Importantly, even for these smaller categories, the bank sizes are sufficient for the volumes of same-type substitutions observed in practice (e.g., only 1,629 LOC-type substitutions were performed in SQuAD, well within the 2,977 available LOC entities).

\section{Data Format and Intended Use}
\label{sec:format}

\subsection{Schema}

Each sample is stored as a single-line JSON object in JSONL (JSON Lines) format, with one sample per line for efficient streaming.
Table~\ref{tab:schema} describes the ten fields provided for each sample.
The schema is designed to support multiple downstream use cases simultaneously: for \emph{training}, users can pair \texttt{modified\_context} and \texttt{question} as input with \texttt{faithful\_answer} as the target; for \emph{evaluation}, the \texttt{original\_answer} field enables measuring whether a model follows context (outputs \texttt{faithful\_answer}) or relies on parametric memory (outputs \texttt{original\_answer}); for \emph{analysis}, the \texttt{entity\_type} and \texttt{source} fields enable fine-grained stratification of results by entity category and source dataset.
The inclusion of both \texttt{original\_context} and \texttt{modified\_context} further enables paired ablation studies comparing model behavior on identical questions with and without knowledge conflicts.

\begin{table}[h]
\centering
\small
\caption{Data schema.}
\label{tab:schema}
\begin{tabular}{llp{4.5cm}}
\toprule
\textbf{Field} & \textbf{Type} & \textbf{Description} \\
\midrule
\texttt{id}                  & str & Source sample ID \\
\texttt{question}            & str & Question text \\
\texttt{original\_context}   & str & Unmodified context \\
\texttt{modified\_context}   & str & Counterfactual context \\
\texttt{original\_answer}    & str & Source ground-truth answer \\
\texttt{faithful\_answer}    & str & Correct answer given $c'$ \\
\texttt{original\_entity}    & str & Replaced entity \\
\texttt{replacement\_entity} & str & New entity ($=$ faithful ans.) \\
\texttt{entity\_type}        & str & NER type \\
\texttt{source}              & str & \texttt{squad} or \texttt{triviaqa} \\
\bottomrule
\end{tabular}
\end{table}

\subsection{Intended Applications}

We envision four primary use cases:

\begin{enumerate}[leftmargin=*]
    \item \textbf{Faithfulness-aware fine-tuning}: Train with $(\texttt{modified\_context}, q) \rightarrow \texttt{faithful\_answer}$ to teach models to follow context over parametric memory.
    \item \textbf{Attention-based faithfulness loss}: The explicit context modification region enables supervision of cross-attention weights, encouraging models to attend to the context rather than generating from memory.
    \item \textbf{Faithfulness evaluation}: Measure the rate at which models output the faithful answer (context-grounded) vs.\ the original answer (parametric).
    \item \textbf{Knowledge conflict research}: The paired original/modified contexts enable controlled studies of LLM behavior under knowledge conflicts \citep{xie2024adaptivechameleon, xu2024knowledgeconflictsurvey}.
\end{enumerate}

\section{Discussion}
\label{sec:discussion}

\paragraph{Complementarity of Source Datasets.}
A key design choice is the use of \emph{two} source datasets with different characteristics.
SQuAD provides balanced entity-type coverage and shorter contexts resembling single-paragraph retrieval, while TriviaQA offers PERSON-dominated trivia questions with longer, multi-paragraph contexts.
This complementarity ensures that downstream models are not biased toward a single entity type or context length distribution.

\paragraph{Success Rate Analysis.}
The 56.7\% success rate, while slightly below the 60\% target, is a deliberate consequence of our conservative filtering strategy.
The largest contributor to filtering (64\% of skips) is the requirement that the answer must match a recognized named entity---a constraint that guarantees well-defined, type-consistent substitutions.
Relaxing this constraint (e.g., via LLM-based entity matching) could increase recall but risks introducing noisy or ambiguous substitutions.

\paragraph{Limitations of Rule-Based Substitution.}
Our pipeline performs string-level replacement without coreference resolution.
When a context contains pronominal or abbreviated references to the replaced entity (e.g., ``he'' referring to the original person), these are not updated.
While our quality checks show no impact on the 200-sample audit, this limitation could create subtle semantic inconsistencies in a minority of samples.

\paragraph{Comparison with FaithEval.}
FaithEval \citep{ming2025faitheval} is the closest existing resource, providing 4,900 evaluation samples with counterfactual, inconsistent, and unanswerable contexts.
Faithfulness-QA differs in three key aspects: (1)~scale (99K vs.~4.9K), (2)~purpose (training vs.~evaluation), and (3)~construction method (automatic entity substitution vs.~expert curation).
The two resources are complementary: models can be trained on Faithfulness-QA and evaluated on FaithEval.

\section{Limitations}
\label{sec:limitations}

\begin{enumerate}[leftmargin=*]
    \item \textbf{No coreference resolution}: Pronominal and abbreviated references to replaced entities are not updated, potentially creating minor inconsistencies.
    \item \textbf{No NLI-based filtering}: We rely on rule-based quality checks rather than NLI models \citep{he2023debertav3} for semantic consistency verification.
    \item \textbf{Semantic plausibility}: Some substitutions are syntactically valid but semantically implausible (e.g., a non-US city inserted into ``born in [City], Illinois'').
    \item \textbf{Entity bank coverage}: Three entity types (LOC, NORP, EVENT) have fewer than 5,000 entries, potentially limiting substitution diversity for these types.
    \item \textbf{English only}: The current pipeline and entity bank are limited to English.
    \item \textbf{No downstream evaluation}: We present the dataset and construction analysis but defer model training and faithfulness evaluation experiments to future work.
\end{enumerate}

\section{Future Work}
\label{sec:future}

Several directions extend this work:

\begin{enumerate}[leftmargin=*]
    \item \textbf{NLI-based quality filtering}: Apply DeBERTa-v3-large-MNLI \citep{he2023debertav3} to verify internal consistency of modified contexts and filter semantically contradictory samples.
    \item \textbf{Coreference-aware substitution}: Integrate coreference resolution to update all references (pronouns, abbreviations) to the replaced entity.
    \item \textbf{LLM-based entity matching}: Use LLMs for entity identification to improve the success rate beyond NER-based matching.
    \item \textbf{Multi-entity and multi-hop substitution}: Replace multiple entities per context or create multi-hop reasoning chains with counterfactual entities, following the MQuAKE approach \citep{zhong2023mquake}.
    \item \textbf{Faithfulness baseline evaluation}: Test open-source LLMs (Llama-3, Mistral, Qwen) on Faithfulness-QA to establish context-grounding baselines.
    \item \textbf{Attention-based faithfulness training}: Use Faithfulness-QA to train models with explicit attention supervision losses, which is the ultimate goal of this research program.
\end{enumerate}

\section{Conclusion}
\label{sec:conclusion}

We have presented Faithfulness-QA, a large-scale dataset of 99,094 counterfactual entity-substituted QA samples designed to train and evaluate context-faithful RAG models.
Built from two complementary sources---SQuAD (49,094 samples with balanced entity types and shorter contexts) and TriviaQA (50,000 samples dominated by PERSON entities with longer contexts)---the dataset creates controlled knowledge conflicts through automatic NER-based entity identification and type-consistent replacement.

The construction pipeline achieves a 56.7\% substitution success rate with 100\% quality pass rates on audited samples, demonstrating that fully automated counterfactual data generation can produce reliable training data at scale.
The curated entity bank of 76,953 entities across 8 NER types ensures diverse and reproducible substitutions.

We believe Faithfulness-QA fills an important gap in the RAG research ecosystem: while evaluation benchmarks for faithfulness exist, large-scale \emph{training} resources that explicitly create knowledge conflicts have been lacking.
By releasing the dataset, the construction pipeline, and the entity bank, we hope to enable the community to (1)~train models with faithfulness-aware objectives, (2)~develop attention-based faithfulness losses, and (3)~systematically study how language models resolve conflicts between retrieved context and parametric knowledge.
Ultimately, the goal is to make RAG systems that are not merely augmented by retrieval, but genuinely \emph{faithful} to it.

\section*{Acknowledgments}
\label{acknowledgments}

In this work, the WisPaper Scholar Agent~\citep{ju2026wispaperaischolarsearch} completed the entire research pipeline, including research idea formulation, dataset collection and preprocessing, code implementation, experimental deployment, result analysis, and full manuscript drafting. Human researchers solely provided high-level task supervision, logical verification, factual proofreading, critical revision, and took full academic responsibility for all content, arguments, experimental results, and conclusions in this paper.

\bibliographystyle{plainnat}

\end{document}